\documentclass[10pt,twocolumn,letterpaper]{article}

\usepackage{cvpr}

\usepackage[pagebackref,breaklinks,colorlinks,allcolors=cvprblue]{hyperref}
\usepackage{booktabs}
\usepackage{xcolor}         
\usepackage{caption}
\usepackage{wrapfig}
\usepackage{amsthm,amssymb,amsmath,bm}
\usepackage{graphicx}
\usepackage{listings}
\usepackage{multirow}
\usepackage{bbm}
\usepackage{algorithmic}
\usepackage{algorithm}
\usepackage{diagbox}
\usepackage{colortbl}

\usepackage{natbib}
\usepackage{amsfonts}
\usepackage{makecell}

\definecolor{cvprblue}{rgb}{0.21,0.49,0.74}
\definecolor{BackgroundBlue}{RGB}{242,242,255}
\definecolor{BackgroundYellow}{RGB}{255,253,233}
\definecolor{BackgroundWhite}{RGB}{255,255,255}
\usepackage[pagebackref,breaklinks,colorlinks,allcolors=cvprblue]{hyperref}

\title{GRPO-RM: Fine-Tuning Representation Models via GRPO-Driven Reinforcement Learning}
\author{
    Yanchen Xu\textsuperscript{\rm 1,2} \quad
    Ziheng Jiao\textsuperscript{\rm 3} \quad
    Hongyuan Zhang\textsuperscript{\rm 1,4}\footnotemark[1] \quad
    Xuelong Li\textsuperscript{\rm 1}\footnote{: Corresponding authors.}\\ 
    {\small\textsuperscript{\rm 1}Institute of Artificial Intelligence (TeleAI), China Telecom}\\
    {\small\textsuperscript{\rm 2} School of Artificial Intelligence, OPtics and ElectroNics (iOPEN), Northwestern Polytechnical University}\\
    {\small\textsuperscript{\rm 3}HuaWei Technologies Co., Ltd.\quad\quad\textsuperscript{\rm 4}The University of Hong Kong}\\
    {\tt\small yanchenxu.tj@gmail.com,\; jzh9830@163.com,\; hyzhang98@gmail.com,\; xuelong\_li@ieee.org}\\
}
\date{}
\begin{document}
\maketitle

\begin{abstract}
	The Group Relative Policy Optimization (GRPO), a reinforcement learning method used to fine-tune large language models (LLMs), has proved its effectiveness in practical applications such as DeepSeek-R1.
	It raises a question whether GRPO can be generalized to representation learning models.
	In this paper, we propose Group Relative Policy Optimization for Representation Model (GRPO-RM), and investigate the performance of GRPO-like policy in post-training representation models.
	Specifically, our method establishes a predefined output set to functionally replace token sequence sampling in LLMs, thereby generating an output group, which is essential for the probability-driven optimization of GRPO.
	In addition, a specialized reward function is designed to accommodate the properties of representation models.
	Extensive experiments are conducted on various real-world datasets to validate the effectiveness of our proposed method.
\end{abstract}

\section{Introduction}
Representation Learning has been a fundamental and crucial research topic \cite{SimCLR, BYOL, MOCO, DINOv1} in computer vision.
The representation models aim to encode comprehensive visual semantics and generate feature embeddings that are invariant with tasks.
These robust image representations are valid for various downstream tasks, including image classification \cite{CoAtNet, Meta_Pseudo_Labels}, semantic segmentation \cite{SVCNet, CFNet}, instance segmentation \cite{Instance_Segmentation}, action recognition \cite{Action_Recognization}, and autonomous driving systems \cite{Autonomous_Driving}.
Generally, despite its outstanding generalization performance, an additional fine-tuning phase is needed to perform downstream tasks, which can also be regarded as the post-training set in Large Language Models \cite{QWEN, LLaMA2}.
Specifically, a task-specific head network is employed to map the extracted features to downstream task objectives.
The overall network is then trained supervisedly to improve the performance.

Since DeepSeek-R1 \cite{DeepSeek-R1} was published, a new optimization method, namely Group Relative Policy Optimization (GRPO), has become popular \cite{DeepSeekMath, DAPO}.
As a reinforcement learning method, GRPO has demonstrated strong capabilities in fine-tuning Large Language Models (LLMs), which naturally leads to a question:
\textbf{\textit{Can GRPO be generalized to representation learning models?}}
In this paper, we are going to answer the above question and develop a GRPO adaption for representation models.

The backbones of representation models vary in different methods.
ResNet \cite{ResNet}, U-Net \cite{UNet}, and Vision Transformers (ViTs) \cite{ViT} are representative network architectures of visual models.
Given that both LLMs and ViTs predominantly utilize transformer architectures, they are likely to exhibit inherent architectural similarities.
Therefore, we attempt to adapt GRPO for post-training on ViT-based representation models.
In this paper, we employ DINOv2 \cite{DINOv2}, a well-known self-supervised ViT model that can generate robust features for various downstream tasks, as the pre-trained base model.

Specifically, we post-train DINOv2 for two representative downstream tasks: image classification and semantic segmentation.
As a cornerstone computer vision task, image classification provides an effective benchmark for assessing a model's ability to extract and leverage global semantic representations.
Semantic segmentation, on the other hand, provides a direct evaluation of local representation quality.
These two tasks align precisely with the DINOv2 output structure, i.e., class tokens and patch tokens.

Our contributions can be summarized as follows:
\begin{itemize}
	\item Based on the properties of representation learning, we redesign the reward functions in the objective of GRPO.
	To be specific, we employ accuracy rewards and uniformity rewards, encouraging the model to generate correct predictions while adaptively discouraging the model to generate wrong ones.
	\item With the novel reward functions, we propose GRPO-RM to post-train DINOv2, which is the first reinforcement post-training method for representation learning models to our best knowledge.
	\item Extensive experiments are conducted on various datasets for different downstream tasks to validate the effectiveness of our method.
	As a result, our method significantly outperforms the standard fine-tuning method.
	In particular, GRPO-RM achieves an average 4.26\% accuracy improvement in out-of-distribution datasets.
\end{itemize}

\section{Related Work}
\subsection{Visual Representation Learning}
Representation learning (RL), which focuses on deriving robust feature embeddings from raw data, is widely explored in the field of computer vision \cite{SimCLR, BYOL, SimSiam, CLIP, DINOv2}.
With techniques like contrastive loss \cite{InfoNCE} and self-distillation \cite{DINOv1}, RL methods manage to eliminate the reliance on labels and demonstrate strong generalization.

Generally, the extracted representations are then fed to a supervised predictor for downstream tasks.
In this paper, we attempt to use a predictor that shares the same architecture but employs a different objective.
Specifically, we replace the standard cross-entropy loss with a novel reinforcement-learning-inspired objective function to optimize representation models.

Moreover, DINOv2 \cite{DINOv2}, which presents strong capability to extract features, is adopted as the pre-trained base model, as mentioned in Section 1.

\subsection{Downstream Tasks of Visual Representation Learning}
Computer vision, serving as a foundational interface for intelligent systems to interpret visual data, plays a crucial role in artificial intelligence.
Therefore, the semantic understanding extracted through the RL models is applicable in various types of downstream tasks, such as image and video classification \cite{Video_Classification}, semantic segmentation \cite{DenseASPP}, and objection detection \cite{Mask_R-CNN, DETR}.

In this paper, we investigate the performance of our proposed method in image classification and semantic segmentation.
Generally, the pre-trained models are fine-tuned with a full-connected layer-based neural network and a Softmax layer for classification.
While for segmentation, we follow the setting of DINOv2, projecting and upsampling the patch tokens to obtain a pixel-level probabilistic distribution.

\subsection{Reinforcement Learning Methods for Post-Training}
Reinforcement learning methods are widely used in the post-training of LLMs to align the model with its users \cite{RLHF, RLAIF}.
Inspired by the idea of TRPO \cite{TRPO}, Proximal Policy Optimization (PPO) \cite{PPO}, which can be implemented much simpler and is more general, is proposed to post-train LLMs based on reinforcement learning.
Another policy that is widely used in post-training LLMs, Direct Preference Optimization (DPO), directly optimizes the model with a classification objective to obtain the best policy that satisfy the preferences, thus avoiding an explicit reward function \cite{DPO}.

In recent years, Group Relative Policy Optimization, namely GRPO, was proposed as a reinforcement learning method to post-train DeepSeek-R1 \cite{DeepSeek-R1}, and has demonstrated exceptional performance in practical applications.
Compared with PPO, GRPO samples multiple outputs for each input question, and compute the group-wise advantages to optimize the policy model.
Without the generalized advantage estimation in PPO, GRPO is able to reduces computational overhead during the training.
Moreover, similar to GRPO, Decoupled Clip and Dynamic sAmpling Policy Optimization (DAPO), an open-source model that dynamically samples output prompts, also achieve success in post-training LLMs \cite{DAPO}.

In this paper, we introduce GRPO-RM, extending GRPO to the post-training of representation model.
This generalization faces some core challenges stemming from architectural differences.
To be specific, the outputs of LLMs are different from those of representation models formally.
Moreover, the advantage computation of GRPO relies on token-level reasoning traces, which lack equivalents in visual feature extractors.
Inspired by the idea of positive-incentive noise \cite{pi-noise,VPN,PiNDA,PiNGDA}, we design the reward functions based on the probabilistic distribution, regarding the probability of wrong outputs as beneficial noise. According to some recent works \cite{PiNI,MIN,RN,Laytrol,MuNG}, proper use of noise can help fine-tuning model, which provides empirical support for our motivation.
\subsection{Downstream Tasks Based on DINO}
As one of the state-of-the-art representation learning methods, DINO is widely employed for all kinds of downstream tasks, which also indicates its generalizability.
Grounding DINO \cite{Grounding-DINO}, which combine DINO with grounded pre-training, aims to detect arbitrary objects, i.e. open-set object detection.
dino.txt \cite{DINO-txt} employs DINO to align self-supervised visual features with language, and achieves great success in zero-shot classification and open-vocabulary semantic segmentation.
Moreover, DINOv2 is also extended to the classification or segmentation of medical images and image generation as well, and performs well in public datasets \cite{Medical-DINO, Medical-DINO-seg, NFIG}.
In this paper, we aim to post-train DINOv2 in two representative CV tasks, showing the advantages of our method over the standard fine-tuning method.
\begin{figure*}[!t]
	\centering
	\includegraphics[width=\textwidth]{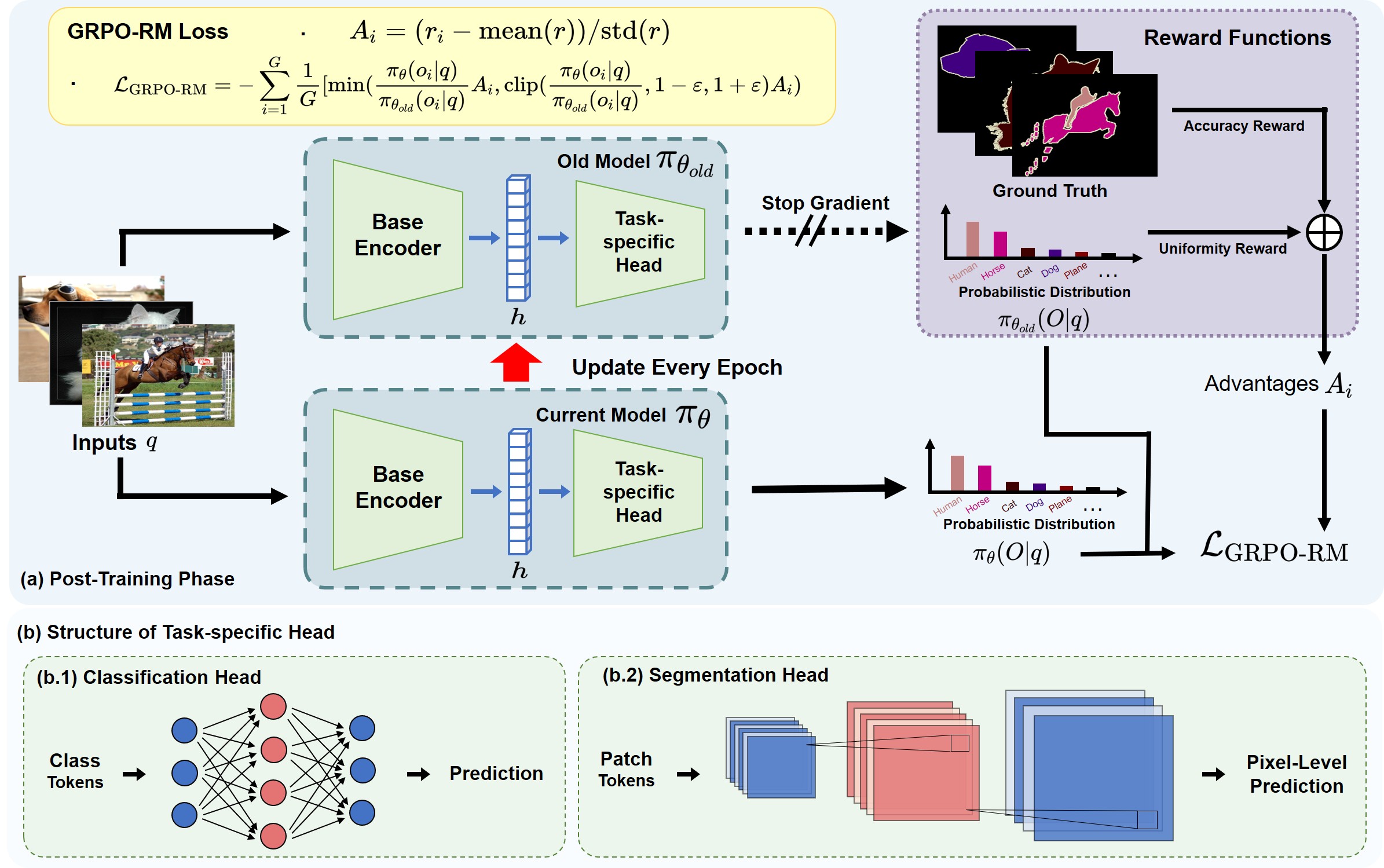}
	\caption{Framework of GRPO-RM:
		(a) The post-training architecture of GRPO-RM comprises a base encoder and a task-invariant head.
		At each epoch, the parameters of the old model ($\theta_{old}$) are updated and incorporated into the loss computation without gradient propagation.
		Advantages are subsequently computed using ground-truth annotations and the probabilistic distributions generated by the old model.
		Finally, the loss is derived from the opposite number of Eq. (\ref{GRPO}) with hyper-parameter $\beta$ fixed to 0.
		(b) Specifically, the network of task-specific heads and the tokens used for post-training vary in different tasks.
		For image classification, we simply feed the class tokens to a full-connected layer-based neural network.
		For semantic segmentation, the patch tokens are upsampled and projected to obtain a pixel-level prediction.}
	\label{Framework}
\end{figure*}

\section{The Proposed Method}
In this section, we first introduce the GRPO adaptation for the post-training of representation models.
Then, with a specially designed reward function, the GRPO-RM method is elaborated.
The framework of our method is illustrated in Figure \ref{Framework}.
\subsection{Preliminaries}
For a question $q$ and a group of corresponding outputs $\{o_1,o_2,\cdots,o_G\}$, the GRPO method with hyper-parameters $\epsilon$ and $\beta$ aims to maximize the objective as follows:
\begin{equation}
	\begin{aligned}
		&\mathcal{J}_{GRPO}(\theta) = \mathbb{E}[q\sim P(Q),\{o_i\}^G_{i=1}\sim \pi_{\theta_{old}}(O|q)]c\\
		&-\beta\mathbb{D}_{KL}(\pi_\theta||\pi_{ref})],
	\end{aligned}\label{GRPO}
\end{equation}
where $\pi_\theta$, $\pi_{\theta_{old}}$, and $\pi_{ref}$ represent the policy model, the old policy model, and the reference model, respectively, and $A_i$ indicate the advantages calculated from the rewards.
As mentioned in Section 2.3, to generalize the GRPO method to post-training of representation models, there are some issues to be addressed:
\begin{itemize}
	\item[$\mathcal{I}1$] Unlike LLMs that support probabilistic sampling of diverse outputs, representation models produce deterministic embeddings. Consequently, the explicit probability distribution (e.g. $\pi_\theta(o_i|q)$ and $\pi_{\theta_{old}}(o_i|q)$) in the objective is not available.
	\item[$\mathcal{I}2$] The reward functions of GRPO are designed for LLMs. It aims to improve accuracy as well as enforce the model to put a thinking process between ``$<$think$>$" and ``$<$/think$>$" tags \cite{DeepSeek-R1}.
	However, this mechanism is incompatible with representation models.
	\item[$\mathcal{I}3$] There may not be a reference model. Hence, the KL divergence term is not available.
\end{itemize}
$\mathcal{I}1$ and $\mathcal{I}2$ will be addressed in Section 3.2 and Section 3.3, respectively.
For $\mathcal{I}3$, we remove the KL divergence term from the objective, which is equivalent to setting hyperparameter $\beta$ to 0.

\subsection{Outputs Sampling for Single Input}
As mentioned in $\mathcal{I}1$, an outputs group with explicit probabilistic distribution is needed in the objectives.
Moreover, the outputs group $\{o_1, o_2,\cdots,o_G\}$ also directly informs the computation of advantages $A_i$, which is the key idea of GRPO.
To encompass the two properties, the output-sampling scheme requires the model to generate different outputs with probabilistic distribution for a single input.
Moreover, the outputs should also support the computation of rewards.
For LLMs, the token sequences meet both needs well, while the outputs of representation models are not.
Intuitively, alternatives for the ``question" and various ``responses" (i.e., token sequences) of LLMs are essential to generalize GRPO method.

Considering that the post-training of representation models is relative to the downstream tasks, the task itself can serve as the questions, take image classification as an example.
The input image can be treated as a question asking about its class, and the whole set of categories can be considered the set of all possible responses.
Thus, a simple Softmax layer yields the probabilistic distribution, and a definite `response' set facilitates the estimation of advantages.

Inspired by this insight, our method fixes the ``response" set and employs a Softmax layer to generate probabilistic distributions over candidates, thereby converting the open-ended Q\&A into a choice selection task.
\subsection{Advantages for Representation Learning}
As mentioned in $\mathcal{I}2$, the advantages in GRPO is incompatible with representation learning models.
To be specific, the advantages $A_i$ in Eq. (\ref{GRPO}) is computed using the rewards $\{r_1,r_2,\cdots,r_G\}$ corresponding to the group of outputs:
\begin{equation}
	A_i = \frac{r_i-\text{mean}(\{r_1,r_2,\cdots,r_G\})}{\text{std}(\{r_1,r_2,\cdots,r_G\})}.\label{advantages}
\end{equation}
However, the reward functions of GRPO are limited to the post-training for LLMs.
To adapt the rewards to representation models, the functions should be redesigned to accommodate their properties.
Intuitively, we require that representations of similar inputs cluster closely in the embedding space, while the overall distribution maintains near-uniform coverage throughout the representation space, i.e., alignment and uniformity \cite{CLModel}.

Inspired by this, we decompose the reward function into two components:
\paragraph{Accuracy Rewards}\quad\\
Similar to the original GRPO methodology, accuracy rewards are adopted, directly encouraging the model to generate correct results.
As the possible output set is fixed, the accuracy reward can be obtained through the corresponding target of the input.

For example, for a dataset with $c$ classes $\{C_1, C_2,\cdots,C_c\}$, given an input $x_k$ of class $C_k$, a group of outputs that exactly stand for $c$ classes will be sampled.
The accuracy reward for each output can be formulated as
\begin{equation}
	\left\{\begin{aligned}
		r_{\text{acc}_i} = 0 &\quad\quad i\neq k,\\
		r_{\text{acc}_i} = c &\quad\quad i=k.
	\end{aligned}\right.\label{AccReward}
\end{equation}
It is worth noting that the value of $r_{{\text{acc}}_k}$ is set to $c$, ensuring that $\mathbb{E}r_{\text{acc}_i}$ is equal to 1 for simplification of calculation.

\paragraph{Uniformity Rewards}\quad\\
Given input $x_k$, the sampled output group $\{o_1,o_2,\cdots,o_c\}$ is corresponding to the $c$ classes of the dataset.
Moreover, a probabilistic distribution $\{p_1, p_2,\cdots, p_c\}$ is obtained through a Softmax layer.
The uniformity rewards are designed to suppress the probability of wrong outputs.
Specifically, it can be formulated as
\begin{equation}
	r_{\text{uni}_i} = -p_i, \quad\quad j\in\{1,2,\cdots,c\}.\label{UniReward}
\end{equation}
Intuitively, for an output that is higher possible to be generated, its uniformity reward will be lower.
Thus, the model is encouraged to output a more uniform probabilistic distribution during the post-training.

Note that the reward of $o_k$, which is the correct output, is also negative to its corresponding probability, which may discourage the model to generate correct output.
As a contrast, another uniformity reward function scheme is as follows:
\begin{equation}
	\left\{\begin{aligned}
		r'_{\text{uni}_i} &= \frac{1-p_k}{c-1}-p_i &\quad\quad i\neq k,\\
		r'_{\text{uni}_i} &= p_i &\quad\quad i=k.
	\end{aligned}\right.\label{OldUniReward}
\end{equation}
In this case, $\sum_{i=1}^cr'_{\text{uni}_i}=1-p_k-\sum_{i\neq k}p_i+p_k=-p_k<1$, ensuring that the accuracy rewards are dominant in the total rewards.
Furthermore, Eq. (\ref{OldUniReward}) encourages the uniformity of the probability of all wrong outputs and is adaptive to each output.
However, there are some issues remaining:
\begin{itemize}
	\item To realize the function in Eq. (\ref{OldUniReward}), more computations are needed because of its asymmetry of different indexes.
	\item In Eq. (\ref{UniReward}), the reward of $o_k\in(-1,0)$, which is far smaller than the corresponding reward in \ref{AccReward}. Hence, there is no need to specifically modify the reward of $o_k$.
\end{itemize}
Based on the above concerns, we employ Eq. (\ref{UniReward}) as the uniformity rewards in GRPO-RM.
Detailed comparisons between Eq. (\ref{UniReward}) and Eq. (\ref{OldUniReward}) will be shown in Section 4.6.

After both components are determined, the final rewards is obtained through addition, i.e.,
\begin{equation}
	r_i = r_{\text{acc}_i} + r_{\text{uni}_i}.\label{Rewards}
\end{equation}

\paragraph{Additional Punishment for Segmentation}\quad\\
It is worth noting that in segmentation datasets, quite a lot of pixels belong to class 1, which stands for the background.
To deal with the imbalanced problem, an additional punishment is applied to the class 1. For each rewards $\{r_1, r_2, \cdots, r_c\}$ obtained through Eq. (\ref{Rewards}), $r_1$ is replaced by $r_1 - c/2$, which is simple for calculation and effectively reduces the impact of the imbalance.
\begin{algorithm}[t]
	\caption{Pseudo code of GRPO-RM}
	\label{Algorithm 1}
	\textbf{Input}: Dataset $\mathcal{D}$, batch size $m$, train epoch $e$, pretrained model $f_\text{pre}$.\\
	\textbf{Hyper-parameters}: $\beta=0,\varepsilon$
	\begin{algorithmic}[1]
		\STATE build the post-train model $\pi_\theta$ with $f_\text{pre}$ and a projection head.
		\FOR{$i=$1 to $e$}
		\STATE $\pi_{\theta_{old}}\leftarrow\pi_\theta$ \textcolor{cvprblue}{$//$ copy old model}
		\FOR{each sampled batch $\{x_i,y_i\}_{i=1}^{m}$}
		\STATE Generate probabilistic distribution: $\pi_{\theta_{old}}(y|x)$.
		\STATE Compute Rewards according to Eq. (\ref{AccReward}), Eq. (\ref{UniReward}), and Eq. (\ref{Rewards}) with $\pi_{\theta_{old}}(y|x)$ and $y_i$.
		\STATE Compute advantages according to Eq. (\ref{advantages}).
		\STATE Compute loss $\mathcal{L}$ according to Eq. (\ref{GRPO}) with advantages, hyper-parameters, $\pi_\theta$, and $\pi_{\theta_{old}}$.
		\STATE Update $\pi_\theta$ to maximize $\mathcal{L}$.
		\ENDFOR
		\ENDFOR
		\STATE \textbf{return} $\pi_\theta$
	\end{algorithmic}
\end{algorithm}
\subsection{Post-Training Models with GRPO-RM}
With all issues mentioned in Section 3.1 addressed, we propose Group Relative Policy Optimization for Representation Model (GRPO-RM), which aims to post-train representation models using reinforcement learning method.

To be specific, GRPO-RM employs the categories of dataset as the output group and the redesigned reward function to compute advantages.
Then, the objective is computed through Eq. (\ref{GRPO}) with hyper-parameter $\beta$ fixed to 0.
The procedure of GRPO-RM is summarized in Algorithm \ref{Algorithm 1}.
Note that the copied model $\pi_{\theta_{old}}$ does not propagate gradient during the training.

\section{Experiments}
In this section, we perform experiments to investigate the following questions:
\begin{itemize}
	\item[\textbf{Q1}] Does GRPO-RM outperform the standard fine-tuning method?
	\item[\textbf{Q2}] What is the time and memory burden of GRPO-RM?
	\item[\textbf{Q3}] How does the uniformity rewards influence the post-training results (as is mentioned in Section 3.3)?
\end{itemize}
\begin{table}[t]
	\centering
	\setlength{\tabcolsep}{1.5mm}
	\caption{Statistics of datasets used in experiments.
		``Classification" stands for image classification, and ``Segmentation" stands for semantic segmentation.}
		\begin{tabular}{cccc}
			\hline\toprule
			\textbf{Dataset} & \textbf{Type} & \textbf{Size} & \textbf{Classes} \\ 
			\midrule
			CIFAR-10 & \multicolumn{1}{|c}{Classification} & 60,000 & 10\\
			CIFAR0-100 & \multicolumn{1}{|c}{Classification} & 60,000 & 100\\
			STL-10 & \multicolumn{1}{|c}{Classification} & 113,000 & 10 \\
			Tiny-ImageNet & \multicolumn{1}{|c}{Classification} & 100,000 & 200 \\
			ImageNet-1k & \multicolumn{1}{|c}{Classification} & 1,331,167 & 1,000 \\
			\midrule
			Pascal VOC 2012 & \multicolumn{1}{|c}{Segmentation} & 1,464 & 21 \\
			ADE20k & \multicolumn{1}{|c}{Segmentation} & 20,210 & 150 \\
			COCO-stuff & \multicolumn{1}{|c}{Segmentation} & 118,287 & 171\\
			\bottomrule\hline
		\end{tabular}
	\label{Datasets}
\end{table}
\subsection{Datasets}
Eight benchmark graph datasets are utilized for experimental study, including five image classification datasets \textbf{CIFAR-10} \cite{CIFAR}, \textbf{CIFAR-100} \cite{CIFAR}, \textbf{STL-10} \cite{STL-10}, \textbf{Tiny-ImageNet} \cite{ImageNet}, and \textbf{ImageNet-1k} \cite{ImageNet} as well as three semantic segmentation datasets \textbf{Pascal VOC} \cite{Pascal-VOC}, \textbf{ADE20k} \cite{ADE20k}, and \textbf{COCO-stuff} \cite{COCO-stuff}.
The details of the datasets are summarized in Table \ref{Datasets}.
\subsection{Experiments settings}
\paragraph{Backbone}\quad\\
Due to the limitation of GPU, all experiments are conducted with a pretrained ViT-S/14 DINOv2 model \cite{DINOv2}.
For image classification datasets, a two-layer fully connected neural network is adopted as the projection head.
For semantic segmentation datasets, we resize the input image size to 448$\times$448 and obtain 32$\times$32 patch tokens.
The tokens are then upsampled to 128$\times$128 as the segmentation map for the segmentation datasets due to the limitation of GPU.
\paragraph{Baselines}\quad\\
We fine-tune the DINOv2 model in a standard way for comparison.
The settings of networks align with those of GRPO-RM.
Moreover, we also conduct experiments on frozen features of the backbone once to directly show the improvement.
\paragraph{Hyper-parameters}\quad\\
For GRPO-RM, the hyper-parameter $\varepsilon$ is set to 0.2.
For experiments on image classification datasets, the dimension of the hidden layer is fixed to 256, and the batch size is 1024 for each dataset.
For experiments on semantic segmentation datasets, the hidden dimension of Pascal VOC is 64 as there are only 21 classes, for the rest datasets, the hidden dimension is also 256.
For all semantic segmentation datasets, the batch size is set to 256.
\begin{table*}[!t]
	\centering
	\setlength{\tabcolsep}{1.5mm}
	\caption{
		Semantic segmentation results (in percent $\pm$ standard deviation) in terms of pixel accuracy, Intersection over Union (IoU), and mean IoU on PASCAL-VOC and ADE20k over 5 runs.
        The setting of feature types is consistent with the setting in \cite{DINOv2}.
		The best result on each dataset is highlighted with \textbf{bold}.}
    \fontsize{9}{10}\selectfont{
		\begin{tabular}{ccccc|ccc}
			\hline\toprule
			\multirow{2}{*}{\textbf{Feature Type}} & \multirow{2}{*}{\textbf{Method}} & \multicolumn{3}{c}{\textbf{Pascal VOC 2012}} & \multicolumn{3}{c}{\textbf{ADE20k}}\\ 
			\cmidrule(lr){3-5}\cmidrule(lr){6-8}
			& & Pixel Acc & IoU & mIoU & Pixel Acc & IoU & mIoU\\
			\midrule
			\multirow{5}{*}{\textbf{Linear}} & \multicolumn{1}{c|}{DINOv2} & \cellcolor{BackgroundWhite} 93.06 & 87.02 & 66.77 & \cellcolor{BackgroundWhite} \textbf{70.19} & 54.07 & 31.52\\
			& \multicolumn{1}{c|}{Fine-tuning (10\%)} & \cellcolor{BackgroundYellow} 93.46 $\pm$ 0.01 & \cellcolor{BackgroundYellow} 88.03 $\pm$ 0.07 & \cellcolor{BackgroundYellow} 69.62 $\pm$ 0.18 & \cellcolor{BackgroundYellow} 69.72 $\pm$ 0.02 & \cellcolor{BackgroundYellow} 54.02 $\pm$ 0.01 & \cellcolor{BackgroundYellow} 32.03 $\pm$ 0.04\\
            & \multicolumn{1}{c|}{Fine-tuning} & \cellcolor{BackgroundYellow} 93.66 $\pm$ 0.04 & \cellcolor{BackgroundYellow} 88.24 $\pm$ 0.02 & \cellcolor{BackgroundYellow} 69.78 $\pm$ 0.11 & \cellcolor{BackgroundYellow} 69.74 $\pm$ 0.02 & \cellcolor{BackgroundYellow} 54.44 $\pm$ 0.02 & \cellcolor{BackgroundYellow} 32.23 $\pm$ 0.03\\
			& \multicolumn{1}{c|}{GRPO-RM (10\%)} & \cellcolor{BackgroundBlue} 93.67 $\pm$ 0.02 & \cellcolor{BackgroundBlue} 88.57 $\pm$ 0.03 & \cellcolor{BackgroundBlue} 70.38 $\pm$ 0.08 & \cellcolor{BackgroundBlue} 69.74 $\pm$ 0.01 & \cellcolor{BackgroundBlue} 54.55 $\pm$ 0.01 & \cellcolor{BackgroundBlue} 32.19 $\pm$ 0.05\\
            & \multicolumn{1}{c|}{GRPO-RM} & \cellcolor{BackgroundBlue} \textbf{93.98 $\pm$ 0.02} & \cellcolor{BackgroundBlue} \textbf{88.64 $\pm$ 0.03} & \cellcolor{BackgroundBlue} \textbf{70.44 $\pm$ 0.12} & \cellcolor{BackgroundBlue} 69.76 $\pm$ 0.01 & \cellcolor{BackgroundBlue} \textbf{55.34 $\pm$ 0.02} & \cellcolor{BackgroundBlue} \textbf{32.62 $\pm$ 0.01}\\
			\midrule
			\multirow{5}{*}{\textbf{Multi Scale}} & \multicolumn{1}{c|}{DINOv2} & \cellcolor{BackgroundWhite} 94.04 & 88.75 & 70.20 & \cellcolor{BackgroundWhite} 71.02 & 55.07 & 32.26\\
            & \multicolumn{1}{c|}{Fine-tuning (10\%)} & \cellcolor{BackgroundYellow} 94.33 $\pm$ 0.01 & \cellcolor{BackgroundYellow} 89.06 $\pm$ 0.05 & \cellcolor{BackgroundYellow} 70.98 $\pm$ 0.04 & \cellcolor{BackgroundYellow} 71.18 $\pm$ 0.02 & \cellcolor{BackgroundYellow} 55.32 $\pm$ 0.01 & \cellcolor{BackgroundYellow} 32.19 $\pm$ 0.05\\
			& \multicolumn{1}{c|}{Fine-tuning} & \cellcolor{BackgroundYellow} 94.34 $\pm$ 0.02 & \cellcolor{BackgroundYellow} 89.19 $\pm$ 0.04 & \cellcolor{BackgroundYellow} 71.45 $\pm$ 0.05 & \cellcolor{BackgroundYellow} 71.29 $\pm$ 0.02 & \cellcolor{BackgroundYellow} 55.80 $\pm$ 0.02 & \cellcolor{BackgroundYellow} 33.19 $\pm$ 0.03\\ 
            & \multicolumn{1}{c|}{GRPO-RM (10\%)} & \cellcolor{BackgroundBlue} \textbf{94.38 $\pm$ 0.01} & \cellcolor{BackgroundBlue} 89.35 $\pm$ 0.01 & \cellcolor{BackgroundBlue} 71.66 $\pm$ 0.06 & \cellcolor{BackgroundBlue} 71.26 $\pm$ 0.01 & \cellcolor{BackgroundBlue} 55.35 $\pm$ 0.01 & \cellcolor{BackgroundBlue} 33.12 $\pm$ 0.02\\
			& \multicolumn{1}{c|}{GRPO-RM} & \cellcolor{BackgroundBlue} \textbf{94.38 $\pm$ 0.01} & \cellcolor{BackgroundBlue} \textbf{89.38 $\pm$ 0.01} & \cellcolor{BackgroundBlue} \textbf{71.71 $\pm$ 0.06} & \cellcolor{BackgroundBlue} \textbf{71.52 $\pm$ 0.01} & \cellcolor{BackgroundBlue} \textbf{56.01 $\pm$ 0.03} & \cellcolor{BackgroundBlue} \textbf{33.58 $\pm$ 0.02}\\
			\bottomrule\hline
		\end{tabular}
		} 
	\label{Semantic_Segmentation}
\end{table*}
\paragraph{Training Settings}\quad\\
For all datasets, an Adam optimizer \cite{Adam} is employed to train the model.
The weight decay factor of the optimizer is fixed to 0.0.
For the learning rate factor in classification tasks, it is initially set to $10^{-3}\times m/256$, where $m$ denotes the batch size.
During the training, the learning rate factor gradually decreases to $10^{-5}\times m/256$.
As for segmentation tasks, the learning rate starts with $10^{-5}\times m/256$ and ends with $10^{-7}\times m/256$
The DINOv2 model is trained for 100 epochs in each dataset.
\paragraph{Evaluation Settings}\quad\\
After post-training, we employ the frozen features of the trained backbone to train a randomly initialized head with an Adam optimizer \cite{Adam} for 50 epochs.
The learning rate and weight decay factor of the optimizer is set in the same way as the training stage.

The experiments are carried out on 4 NVIDIA A100 vGPUs with 40 GB memory or 2 NVIDIA H200 vGPUS with 140 GB memory.
\begin{table}[t]
	\centering
	\setlength{\tabcolsep}{0.8mm}
	\caption{
		Accuracy (in percent $\pm$ standard deviation) on image classification datasets over 5 runs.
		The best result on each dataset is highlighted with \textbf{bold}.}
		\begin{tabular}{cccc}
			\hline\toprule
			\textbf{Dataset} & \textbf{Method} & \textbf{SR} & \textbf{$k$NN}\\
			\midrule
			\multirow{3}{*}{CIFAR-10} & DINOv2  & 47.26 & 43.26\\
			& Fine-tuning & \cellcolor{BackgroundYellow} 94.55 $\pm$ 0.27 & \cellcolor{BackgroundYellow} 94.39 $\pm$ 0.29\\
			& GRPO-RM & \cellcolor{BackgroundBlue} \textbf{95.78 $\pm$ 0.26} & \cellcolor{BackgroundBlue} \textbf{95.81 $\pm$ 0.30}\\
			\midrule
			\multirow{3}{*}{CIFAR-100} & DINOv2 & 20.68 & 25.92\\
			& Fine-tuning & \cellcolor{BackgroundYellow} 68.87 $\pm$ 2.70 & \cellcolor{BackgroundYellow} 71.32 $\pm$ 1.85\\
			& GRPO-RM & \cellcolor{BackgroundBlue} \textbf{74.41 $\pm$ 2.28} & \cellcolor{BackgroundBlue} \textbf{78.48 $\pm$ 0.92}\\
			\midrule
			\multirow{3}{*}{STL-10} & DINOv2 & 15.71 & 37.57\\
			& Fine-tuning & \cellcolor{BackgroundYellow} 43.44 $\pm$ 6.89 & \cellcolor{BackgroundYellow} \textbf{89.76 $\pm$ 3.50}\\
			& GRPO-RM & \cellcolor{BackgroundBlue} \textbf{46.42 $\pm$ 5.86} & \cellcolor{BackgroundBlue} 89.03 $\pm$ 4.43\\
			\midrule
			\multirow{3}{*}{Tiny-ImageNet} & DINOv2 & 29.24 & 21.10\\
			& Fine-tuning & \cellcolor{BackgroundYellow} 58.67 $\pm$ 0.61 & \cellcolor{BackgroundYellow} 58.59 $\pm$ 1.25\\
			& GRPO-RM & \cellcolor{BackgroundBlue} \textbf{65.96 $\pm$ 0.40} & \cellcolor{BackgroundBlue} \textbf{66.36 $\pm$ 0.32}\\
			\midrule
			\multirow{3}{*}{ImageNet} & DINOv2 & 74.34 & \textbf{72.61}\\
			& Fine-tuning & \cellcolor{BackgroundYellow} 74.48 $\pm$ 0.85 & \cellcolor{BackgroundYellow} 71.33 $\pm$ 0.45\\
			& GRPO-RM & \cellcolor{BackgroundBlue} \textbf{76.21 $\pm$ 0.92} & \cellcolor{BackgroundBlue} 72.16 $\pm$ 0.73\\
			
			\bottomrule\hline
		\end{tabular}
	\label{Image_Classification}
\end{table}
\begin{table}[t]
	\centering
	\setlength{\tabcolsep}{1mm}
	\caption{
		Semantic segmentation results in terms of pixel accuracy, Intersection over Union (IoU), and mean IoU on COCO-stuff.
		The best result on each dataset is highlighted with \textbf{bold}.}
    \fontsize{9}{10}\selectfont{
		\begin{tabular}{ccccc}
			\hline\toprule
			\multirow{2}{*}{\textbf{Feature Type}} & \multirow{2}{*}{\textbf{Method}} & \multicolumn{3}{c}{\textbf{COCO-stuff}}\\ 
			\cmidrule(lr){3-5}
			& & Pixel Acc & IoU & mIoU\\
			\midrule
			\multirow{5}{*}{\textbf{Linear}} & \multicolumn{1}{c|}{DINOv2} & \cellcolor{BackgroundWhite} 63.43 & 46.45 & 30.79\\ 
			& \multicolumn{1}{c|}{Fine-tuning (10\%)} & \cellcolor{BackgroundYellow} 63.71 & \cellcolor{BackgroundYellow} 46.67 & \cellcolor{BackgroundYellow} 30.94\\ 
            & \multicolumn{1}{c|}{Fine-tuning} & \cellcolor{BackgroundYellow} 64.08 & \cellcolor{BackgroundYellow} 47.72 & \cellcolor{BackgroundYellow} 31.67\\
			& \multicolumn{1}{c|}{GRPO-RM (10\%)} & \cellcolor{BackgroundBlue} 63.72 & \cellcolor{BackgroundBlue} 47.51 & \cellcolor{BackgroundBlue} 31.07\\
            & \multicolumn{1}{c|}{GRPO-RM} & \cellcolor{BackgroundBlue} \textbf{64.10} & \cellcolor{BackgroundBlue} \textbf{48.01} & \cellcolor{BackgroundBlue} \textbf{32.25}\\
			\midrule
			\multirow{5}{*}{\textbf{Multi Scale}} & \multicolumn{1}{c|}{DINOv2} & \cellcolor{BackgroundWhite} 64.57 & 47.68 & 31.75\\
            & \multicolumn{1}{c|}{Fine-tuning (10\%)} & \cellcolor{BackgroundYellow} 64.87 & \cellcolor{BackgroundYellow} 47.98 & \cellcolor{BackgroundYellow} 32.77\\ 
			& \multicolumn{1}{c|}{Fine-tuning} & \cellcolor{BackgroundYellow} 64.99 & \cellcolor{BackgroundYellow} 48.26 & \cellcolor{BackgroundYellow} 32.88\\
            & \multicolumn{1}{c|}{GRPO-RM (10\%)} & \cellcolor{BackgroundBlue} 64.94 & \cellcolor{BackgroundBlue} 48.08 & \cellcolor{BackgroundBlue} 32.22\\
			& \multicolumn{1}{c|}{GRPO-RM} & \cellcolor{BackgroundBlue} \textbf{65.02} & \cellcolor{BackgroundBlue} \textbf{48.53} & \cellcolor{BackgroundBlue} \textbf{33.06}\\
			\bottomrule\hline
		\end{tabular}
		} 
	\label{Semantic_Segmentation_COCO}
\end{table}

\subsection{Performance Analysis (Q1)}
To evaluate the effectiveness of GRPO-RM, the performance analysis is essential.
Consequently, experiments are conducted on two different, representative downstream tasks, image classification and semantic segmentation.
To better demonstrate the performance, we also visualize the results of semantic segmentation.
\paragraph{Image Classification}\quad\\
For classification accuracy, we measure the performance through Softmax Regression and $k$NN accuracy with $k=5$.
The classification accuracy is reported in Table \ref{Image_Classification}.
Our method achieves an average 3.754\% Softmax Regression (SR) improvement and 3.29\% $k$NN improvement than standard fine-tuning method (denoted by ``Fine-tuning" in table), which is more significant in out-of-distribution datasets (i.e., the datasets expect for ImageNet), with an average 4.26\% SR improvement and 3.905\% $k$NN improvement.
For example, in Tiny-ImageNet, GRPO-RM outperforms the standard fine-tuning method by 7.29\% in SR and 7.77\% in $k$NN.
Moreover, GRPO-RM also has a lower standard deviation in most datasets, demonstrating the stability of our method.
\paragraph{Semantic Segmentation}\quad\\
For segmentation datasets, we measure the performance through three metrics, the pixel accuracy, Intersection over Union (IoU), and mean IoU.
As both approaches take a long time, we conduct further experiments that use a random 10\% of the dataset during the training for comparison, corresponding to a quick and small-scale post-training.
The results of the standard fine-tuning method and GPRO-RM are reported in Table \ref{Semantic_Segmentation} and Table \ref{Semantic_Segmentation_COCO}.
As a result, our method performs better on all datasets.

\begin{figure*}[t]
	\centering
	\includegraphics[width=\textwidth]{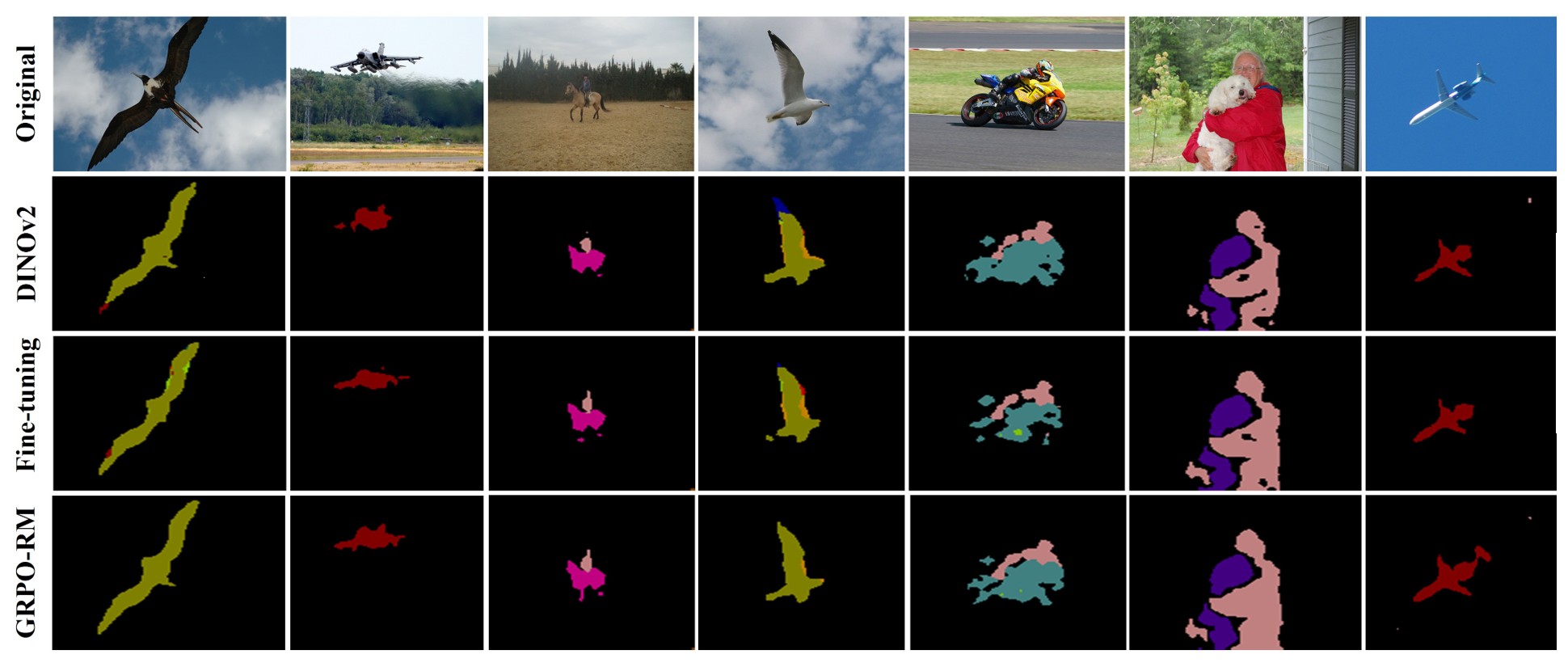}
	\caption{Visualization of DINOv2, Fine-tuning, and GRPO-RM on PASCAL-VOC.}
	\label{Visualize}
\end{figure*}

\paragraph{Visualization}\quad\\
To further demonstrate the effectiveness of GRPO-RM in semantic segmentation tasks, we visualize the segmentation results of DINOv2, standard fine-tuning, and GRPO-RM in Fig \ref{Visualize}.
As can be seen from the visualization, GRPO-RM achieves a more accurate segmentation of the image details.

\begin{figure*}[t]
	\centering
	\includegraphics[width=\textwidth]{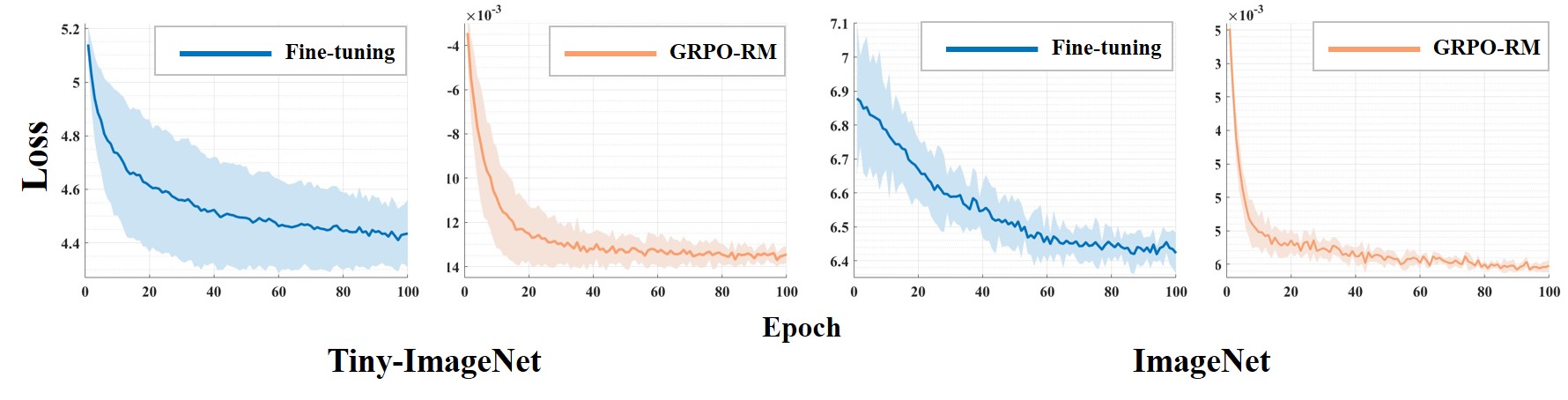}
	\caption{
		Training loss curves for GRPO-RM versus baseline on ImageNet (in-distribution) and Tiny-ImageNet (out-of-distribution).
		It can be easily derived from the figure that GRPO-RM converge much faster than normal post-training method.}
	\label{Convergence}
\end{figure*}
\subsection{Convergence Rate (Q1)}
Considering that the reinforcement learning method is never used to post-train a representation learning model, it is essential to verify whether the proposed GRPO-RM can converge through training.

To be specific, we investigate the convergence rate of GRPO-RM through the loss curves on different datasets, including an out-of-distribution dataset, Tiny-ImageNet and an in-distribution dataset, ImageNet.
We also draw the corresponding curve of the standard fine-tuning method as a comparison.
The results are shown in Fig \ref{Convergence}.

From Fig \ref{Convergence}, it is easy to find that GRPO-RM exhibits rapid loss reduction rate during post-training, achieving near-convergence within 20 epochs
For the standard fine-tuning method, the loss keeps decreasing in terms of overall trends until 100 epochs are finished.
This indicates that GRPO-RM converges much faster.

\subsection{Computational Burden (Q2)}
The computational burden of GRPO-RM is shown in Table \ref{Burden}.
Three metrics are adopted to evaluate the costs, including training time for each epoch, floating point operations (FLOPs), and GPU memory required.

As a result, GRPO-RM requires a bit more computation overhead compared to standard fine-tuning methods, which is an acceptable trade-off given its accelerated convergence and the accuracy improvement as well.
\begin{table}[t]
	\centering
    \setlength{\tabcolsep}{1.2mm}
	\caption{
		Comparison of computational costs in different datasets.
		For classification datasets, the batch size is set to 1024, while for segmentation datasets, the batch size is set to 256.
		Moreover, the patch tokens are upsampled to 128$\times$128.
	}
    \fontsize{9}{10}\selectfont{
		\begin{tabular}{ccccc}
			\hline\toprule
			\textbf{Dataset} & \textbf{Method} & \textbf{Time} & \textbf{FLOPs} & \textbf{Memory}\\ 
			\midrule
			\multirow{2}{*}{\textbf{Tiny-ImageNet}} & Fine-tuning & 9.30s & 5.72e11 & 13.26G\\ 
			& GRPO-RM & 10.37s & 7.15e11 & 15.12G\\
			\cmidrule(lr){2-5}
			\multirow{2}{*}{\textbf{ImageNet}} & Fine-tuning & 175.95s & 5.66e12 & 95.40G\\ 
			& GRPO-RM & 238.20s & 7.07e12 & 107.00G\\
			\midrule
			\multirow{2}{*}{\textbf{Pascal VOC}} & Fine-tuning & 151.39s & 2.90e12 & 217.72G\\ 
			& GRPO-RM & 174.29s & 5.80e12 & 220.66G\\
			\cmidrule(lr){2-5}
			\multirow{2}{*}{\textbf{COCO-stuff}} & Fine-tuning & 25.83min & 3.39e12 & 223.46G\\ 
			& GRPO-RM & 27.83min & 6.78e12 & 232.04G\\
			\bottomrule\hline
		\end{tabular}
	} 
	\label{Burden}
\end{table}
\begin{table}[t]
	\centering
	\caption{
		Comparison in terms of accuracy and training time with different uniformity reward function settings.
		``set to 0" means only an accuracy reward is applied, and the rest two items denotes the reward function with Eq. (\ref{OldUniReward}) and Eq. (\ref{UniReward}) as the uniformity rewards, respectively.
	}
	\setlength{\tabcolsep}{1.5mm}
	\fontsize{9}{10}\selectfont{
		\begin{tabular}{cccc}
			\hline\toprule
			\textbf{Dataset} & \textbf{Uniformity Reward} & \textbf{SR} & \textbf{Time}\\
			\midrule
			\multirow{3}{*}{\textbf{STL-10}} & set to 0 & 34.90 & 1.11s\\
			& computed through Eq. (\ref{OldUniReward}) & 36.27 & 3.56s\\
			& Eq. (\ref{UniReward}) (GRPO-RM) & 46.42 & 0.95s\\
			\cmidrule(lr){2-4}
			\multirow{3}{*}{\makecell[c]{\textbf{Tiny-}\\\textbf{ImageNet}}} & set to 0 & 62.16 & 9.40s\\
			& computed through Eq. (\ref{OldUniReward}) & 60.32 & 18.57s\\
			& Eq. (\ref{UniReward}) (GRPO-RM) & 65.96 & 10.37s\\
			\cmidrule(lr){2-4}
			\multirow{3}{*}{\textbf{ImageNet}} & set to 0 & 77.04 & 232s\\
			& computed through Eq. (\ref{OldUniReward}) & 74.70 & 251.3s\\
			& Eq. (\ref{UniReward}) (GRPO-RM) & 76.21 & 238.2s\\
			\midrule
			\multirow{3}{*}{\makecell[c]{\textbf{Pascal}\\\textbf{VOC 2012}}} & set to 0 & 93.61 & 138.27s\\
			& computed through Eq. (\ref{OldUniReward}) & - & 157min\\
			& Eq. (\ref{UniReward}) (GRPO-RM) & 94.67 & 174.29s\\
			\bottomrule\hline
		\end{tabular}
		} 
	\label{Ablation}
\end{table}
\subsection{Ablation Study (Q3)}
As mentioned in Section 3.3, further experiments are conducted to compare the effect while using Eq. (\ref{UniReward}) or Eq. (\ref{OldUniReward}) as uniformity reward functions.
We also employ the accuracy rewards as the only reward function to compute advantages, as another comparison.
The different uniformity settings are denoted as ``set to 0", ``computed through Eq. (\ref{OldUniReward})", and ``Eq. (\ref{UniReward}) (GRPO-RM)" in Table \ref{Ablation}, respectively.

As a result, when Eq. (\ref{OldUniReward}) is employed as the uniformity reward function, the method shows relatively poor performance in both SR accuracy and training time.
The reason might be that the punishment in Eq. (\ref{OldUniReward}) of wrong output is not strong enough, and may have an encouraging effect in some occasions.
The disadvantages of Eq. (\ref{OldUniReward}) is more significant in segmentation datasets, as the formula requires much more computation, resulting in a much longer training time, which is over 50 times of GRPO-RM.

On the other hand, compared with the method that only employs accuracy reward, GRPO-RM requires an average 6.16\% extra training time while exhibits stronger performance for an average 3.88\% SR accuracy improvement, which demonstrates an optimal efficiency-performance tradeoff.
Moreover, the results also validate the effectiveness of the uniformity rewards, especially for out-of-distribution datasets like STL-10 and Tiny-ImageNet.
\section{Conclusion}
In this paper, we propose a reinforcement post-training framework, namely GRPO-RM, that adapts Group Relative Policy Optimization to representation models, which is the first reinforcement learning method for post-training of the visual representation models, to our best knowledge.
Inspired by the success of GRPO in post-training LLMs, we reconfigure its objective for visual representations.
To be specific, we regard the predictions with probabilities as the output group and design reward functions that align with the properties of representation learning for them, encouraging the model when the probability of correct
predictions is high while discouraging it for the opposite.
Extensive experiments conducted on various datasets with different tasks validate the effectiveness of our method over standard post-training methods.
Although the computational burden experiment shows that GRPO-RM requires more training time and memory, GRPO-RM display a better performance in accuracy and a much faster convergence rate, which is an acceptable trade-off.
We also validate the effectiveness of our designed reward functions through the ablation study.
{
    \small
    \bibliographystyle{ieeenat_fullname}
    \bibliography{citations}
}

\end{document}